\documentclass[conference]{IEEEtran}

\usepackage{graphicx} 
\usepackage{float}
\usepackage{cite}
\usepackage{amsmath,amssymb,amsfonts}
\usepackage{algorithmic}
\usepackage{textcomp}
\usepackage{pgfplots}
\usepackage{booktabs}
\usepackage{subcaption}
\usepackage{comment}
\usepackage{multirow}
\pgfplotsset{compat=1.17}
\usepackage{array}
\usepackage{multicol}
\usepackage{titlesec}
\usepackage{xcolor}
\usepackage[bookmarks=false]{hyperref}
\title{Task-Centric Multimodal Learning for Cognitive Impairment Detection
}
\author{\IEEEauthorblockN{ 
Sankalp Vaish$^1$,
Rishank Singh$^1$,
Sumukh Veeramalla$^1$,
Caroline Summerour$^2$,
John A.  Batsis$^2$,
Xiaohui Liang$^1$
}

		\IEEEauthorblockA{$^1$Department of Computer Science,
	University of Massachusetts Boston, MA, USA
    }
    \IEEEauthorblockA{$^2$School of Medicine,
	University of North Carolina at Chapel hill, NC, USA
		}
		Email: \{Sankalp.Vaish001, Rishank.Singh001, Sumukh.Veeramalla001, Xiaohui.Liang\}@umb.edu, \\Caroline\_Summerour@med.unc.edu, John.Batsis@unc.edu
}
\date{September 2025}

\begin{document}

\maketitle

\begin{center}
\footnotesize
\copyright~2026 IEEE. Personal use of this material
is permitted. Permission from IEEE must be obtained
for all other uses, in any current or future media,
including reprinting/republishing this material for
advertising or promotional purposes, creating new
collective works, for resale or redistribution to
servers or lists, or reuse of any copyrighted
component of this work in other works.

\vspace{0.5em}

\textit{Accepted for publication at the
2026 IEEE Global Communications Conference
(GLOBECOM).}
\end{center}

\begin{abstract}
Dementia affects millions of older adults worldwide, yet early detection remains difficult outside clinical settings. Traditional models 
primarily rely on text or voice-based responses, capturing only a limited view of human cognition despite the inherently multimodal nature of human expression. This work proposes a contactless, multimodal screening framework that leverages video, audio, and text signals collected during structured voice-based interactions with Amazon Alexa. By incorporating these modalities, our approach captures a more holistic representation of human behavior encompassing speech patterns, vocal characteristics, facial expressions, and linguistic structure, which may enable more accurate detection of cognitive decline. 
Different participants completed three conversational tasks designed to probe spontaneous speech, comprehension, and recall.
We pursued three objectives: (1) evaluate which of the three Alexa tasks provides the most informative cognitive signals, (2) compare the effectiveness of unimodal models, and (3) assess the benefits of multimodal fusion models. 
Task-level analysis shows that the video model achieves its best performance on Command-Reading, with an accuracy of 0.65 (AUC 0.69), while the audio model performs best on Question-Answering, reaching an accuracy of 0.67 (AUC 0.73). In contrast, the multimodal model demonstrates consistently stable performance across tasks, achieving an accuracy of 0.76 (AUC 0.75) on Command-Reading and 0.76 (AUC 0.73) on Question-Answering.
These findings demonstrate that multimodal behavioral cues captured through naturalistic interactions can reliably distinguish MCI from healthy aging, highlighting the limitations of traditional diagnostic paradigms and supporting the shift toward multimodal, real-world cognitive assessment frameworks. 
\end{abstract}

\section{Introduction}

Dementia is a growing global health concern, particularly among older adults, and its prevalence continues to increase with aging populations. Early detection is crucial, as timely intervention can significantly improve quality of life and inform effective care planning. Currently, diagnosis is heavily based on standardized clinical assessments, such as the Montreal Cognitive Assessment (MoCA). 
Mild cognitive impairment (MCI)~\cite{alzheimers_mci} is a condition in which people have more memory or thinking problems than other people their age. Individuals with MCI face an elevated risk of progressing to Alzheimer’s disease or related dementias, with studies estimating that approximately one to two out of every ten adults aged 65 or older with MCI develop dementia each year.

Despite the growing success of cognitive impairment~\cite{AudioAn1, RI2} detection, most existing studies primarily focus on improving feature representations or fusion strategies, while treating the underlying interaction context as secondary. This leads to a gap in understanding how different tasks influence the quality and discriminative power of behavioral signals. 
To address this, there is a need for a systematic, task-centric investigation that explicitly analyzes how different interaction paradigms contribute to cognitive assessment.
 In real-world conversational settings, individuals do not produce uniform responses; instead, their speech, language, and behavioral cues vary significantly depending on the cognitive demands imposed by the task. To account for this variability, we design our experimental framework to explicitly model task-dependent behavior. Specifically, instead of treating participant data as homogeneous, we consider three distinct interaction paradigms: Command-Reading (task-1), Command-Generation (task-2), and Question-Answering (task-3), each designed to elicit different cognitive and linguistic processes.

 A substantial amount of prior work has focused on speech-based dementia detection, examining acoustic, prosodic, and linguistic biomarkers extracted from conversational speech~\cite{Speech1,Speech2}. Other studies have extended this line of research by incorporating transcript-derived linguistic cues alongside audio features~\cite{AudioTrans1}. While these methods demonstrate the utility of speech for cognitive assessment, they largely overlook additional behavioral channels. 
 To address this limitation, we incorporate visual behavioral cues by analyzing detailed facial emotion patterns~\cite{Face1} and Action Units~\cite{AU1} derived from video recordings to characterize differences between cognitively healthy and MCI participants.

In this work, we argue that the nature of the collection task itself plays a central role in determining the sensitivity of behavioral signals to early cognitive decline.
We further investigate whether such early signals can be detected through natural facial behavior, speech, and language use. In addition, we analyze how these behavioral cues vary across tasks, providing insight into which interaction paradigm is most informative.
Beyond facial cues, we extract audio biomarkers and textual linguistic features, building on prior speech-driven dementia studies~\cite{Speech1}. Each modality is first evaluated independently, followed by integration within a unified multimodal framework to assess the benefits of fusion over individual signals. Our findings show that such multimodal models can detect early cognitive impairment even from short, casual interactions.
This work makes the following key contributions:

\begin{enumerate}


\item \textbf{A conversational dataset with structured interaction tasks.}
We created a dataset collected across multiple sessions through three distinct Alexa-based tasks. Unlike traditional clinical datasets, these tasks are designed to capture diverse cognitive behaviors, including natural conversation, semi-structured interaction, and memory-based question answering. This diversity enables the analysis of task-specific cognitive signals, providing a more realistic setting for home-based screening.

\item \textbf{Unified multimodal preprocessing pipeline:}  
We develop an end-to-end pipeline for extracting standardized features across all three modalities, including (i) facial Action Units and frame-level emotion probabilities from video, (ii) acoustic and prosodic biomarkers from participant speech, and (iii) linguistic, semantic, and cognitive-linguistic features from transcripts. These then act as an input to a transformer encoder architecture with multi-head self-attention to capture time-series data.

\item \textbf{Task-level analysis:}
We observe that unimodal models are highly task-sensitive across the three interaction paradigms. The video model performs better on task-1 and task-2, achieving an accuracy of 0.65 (AUC 0.69) on Task 1, but degrades on Task 3, where it attains an accuracy of 0.59 (AUC 0.59). In contrast, the audio-based model performs best on task-2 and task-3, reaching its highest performance on task-3 with an accuracy of 0.67 (AUC 0.73).



\item \textbf{Comprehensive evaluation of unimodal and multimodal models:}  
We treat each task as a distinct data point and observe consistent trends across modalities. The video model produces a higher number of Type I errors (41) compared to Type II errors (22), whereas the audio model exhibits the opposite behavior, with more Type II errors (46) than Type I errors (14). In contrast, the multimodal model reduces both error types to 21 (Type I) and 16 (Type II), 
indicating that multimodal fusion effectively leverages complementary predictive strengths of different modalities.


\end{enumerate}

\section{Related Work}

 Early detection of dementia traditionally relies on standardized clinical assessments, which require trained examiners and controlled environments, limiting scalability and frequency of testing. As a result, automated and contactless approaches have gained significant attention.
Facial behavior, like emotional expressions and action units, offers another rich modality for cognitive health analysis. Prior studies have demonstrated that facial expressions may help differentiate individuals with dementia from healthy controls~\cite{Face1}. However, such work typically treats facial cues as a stand-alone modality, without integrating them with speech or linguistic features. To date, no existing study has combined audio, text, and video into a unified multimodal framework for dementia detection, despite increasing evidence that cognitive decline manifests across all three domains.
Recent advances in affective computing and sentiment analysis further highlight the value of multimodal behavioral cues in machine learning systems~\cite{Sent1,Sent2}.
Building on this, Large Language Models (LLMs) have rapidly emerged in healthcare applications, supporting clinical decision-making, diagnostics, and multimodal reasoning. These systems leverage domain-specific pretraining, fine-tuning, and prompt engineering techniques to interpret diverse data types~\cite{LLMH1
}. Despite this progress, the potential of LLMs to reason over multimodal behavioral biomarkers for cognitive impairment remains largely unexplored.
Some research has also incorporated neuroimaging modalities such as MRI and PET scans for dementia diagnosis~\cite{brain1,brain2}. Although effective, these methods are costly, inaccessible, and unsuitable for home-based or longitudinal monitoring. In contrast, behavioral signals from everyday interactions (speech, facial expressions, and language) offer a scalable alternative that can be captured unobtrusively through common consumer devices.


A recent study, Cog-TiPRO~\cite{CogTiPRO}, demonstrated the feasibility of using voice-assistant-based speech and linguistic data for longitudinal detection of cognitive decline. However, that work, like most prior research~\cite{RI3, RI2}, omits video-derived behavioral cues. Our work advances this direction in three important ways: (1) by adding facial expression and Action Unit analysis, capturing non-verbal indicators that may manifest earlier than speech changes; (2) by combining video, audio, and text in a unified multimodal pipeline; and (3) by evaluating across multiple structured interaction types and sessions, enabling a more comprehensive and ecologically valid evaluation for real-world, home-based screening.

\section{Methodology}

\subsection{Preprocessing}

\begin{figure}[ht]
    \centering
    \includegraphics[width=1\linewidth]{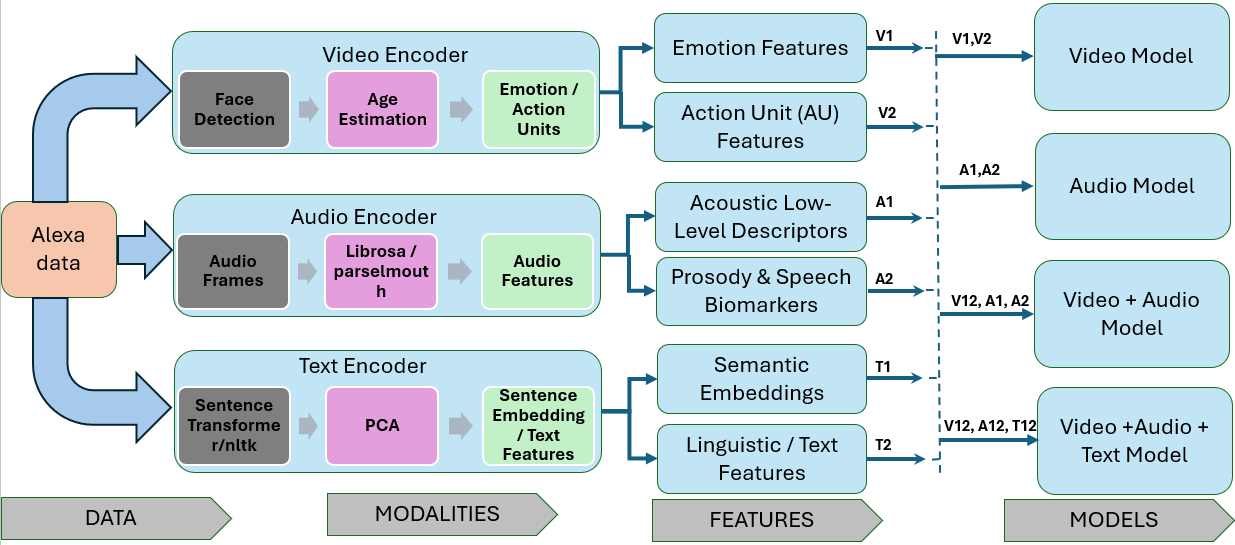}
    \caption{Processing Pipeline}
    \label{fig: Processing Pipeline}
\end{figure}

The dataset consisted of synchronized audio–video recordings of participant interactions with Alexa, complemented by automatic transcripts generated via Rev.ai. All three modalities were preprocessed to extract meaningful behavioral features before their integration into the unimodal and multimodal models. Figure~\ref{fig: Processing Pipeline} illustrates the pipeline from raw recordings to modality specific feature embeddings and
their mapping to various modeling architectures. 

\textbf{Video Encoder:}
To address the presence of multiple individuals in each recording (participant, interviewer, Alexa device, questionnaire holder), we first applied MiVOLO~\cite{MiVOLO} to every frame for face detection and age estimation. This allowed us to isolate the older adult participant across all sessions. 
\textit{Emotion Features:} We employed ResEmoteNet~\cite{ResEmoteNet} to estimate frame-level emotional states (happy, sad, angry, surprise, neutral, disgust, fear). For each speaking segment, we aggregated these predictions into quantitative descriptors~\cite{Face1}, including:
emotion counts, emotion distribution, emotion variability, and frequency of emotion shifts.
\textit{Action Unit Features:}
To capture more granular facial behavior, we used OpenFace~\cite{OpenFace1}
to extract Facial Action Units (AUs), which quantify the activation of specific facial muscles (e.g., AU1: Inner Brow Raise, AU12: Lip Corner Pull). These AUs provide a physiologically grounded representation of subtle facial movements ~\cite{AU1} that may correlate with cognitive decline. 

\textbf{Audio Encoder}
Using Parselmouth and Librosa, we analyzed only the participant’s speech segments~\cite{AudioAn1,AudioAn2}. Two families of features were extracted:
\textit{Acoustic Features:} MFCC statistics, Spectral centroid, Spectral flux, Harmonics-to-noise ratio.
\textit{Prosodic and Speech Biomarkers:} Pitch (F0) metrics, Intensity contours, Jitter and shimmer, Speaking rate, Pause and Articulation rate.

\textbf{Text Encoder:}
Transcripts generated by Rev.ai were processed using NLTK, TextBlob, and Sentence Transformers~\cite{text1,text2}
to derive these features:
\textit{Linguistic and Semantic Features: } Lexical richness, Syntactic complexity, Semantic coherence, Use of filler words, hesitations, and discourse markers.
\textit{Semantic Embeddings}
Sentence-level embeddings were extracted using a Sentence Transformer model, followed by Principal Component Analysis (PCA) to retain the top 10 components representing the semantic structure of speech.

\subsection{Modeling}

Our modeling framework systematically evaluates the contribution of different tasks using a diverse set of models, ranging from traditional machine learning classifiers to multimodal transformer architectures that capture temporal dynamics in the extracted features from the above-described pipeline. We use a standard transformer encoder architecture with multi-head self-attention for sequence modeling~\cite{TS1},~\cite{TS2,RI}.
We first trained separate models on each of the three tasks to identify which task provides the most informative signals for dementia detection. We then evaluated unimodal models for each task and compared their performance with a multimodal model that integrates information across tasks and modalities, enabling us to isolate the cognitive and behavioral cues.

\section{Dataset}

The dataset was collected from 35 older adult participants through a series of structured interactions with Amazon Alexa. Each experimental session was divided into three distinct tasks, designed to elicit different forms of verbal 
engagement:

Command-Reading (task-1): Participants were provided with a list of 30 predefined questions (e.g., “What is the weather outside?”, “What is today’s date?”) and instructed to ask Alexa directly. This task primarily assessed verbal clarity and recall when following straightforward instructions.
Command-Generation (task-2): Participants were given 30 intent prompts (e.g., weather, date, news) and asked to independently formulate appropriate questions to pose to Alexa. This task introduced a higher cognitive load, evaluating language generation, flexibility, and planning. Question-Answering (task-3): Alexa presented participants with 30 pre-recorded questions (e.g., “Name some animals you would find in a zoo”), and participants were required to respond verbally. This task targeted comprehension, memory recall, and spontaneous speech production.

\begin{figure}[ht]
    \centering
    \includegraphics[width=1\linewidth]{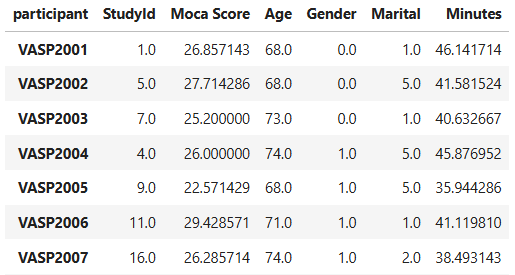}
    \caption{Data}
    \label{fig: Data}
\end{figure}

Figure \ref{fig: Data} provides an overview of the dataset structure, illustrating the duration and composition of each experimental session. One session lasted approximately 35–75 minutes and included all three tasks and was repeated 7 times over a period of 18 months. The figure also displays the MoCA score, which serves as the primary ground-truth label for cognitive status estimation. Participants with MoCA scores between 18 and 25 were categorized as having Mild Cognitive Impairment (MCI), while those scoring 26 or above were considered cognitively healthy.
Finally, the figure lists all participant IDs and their corresponding session numbers. No personal or identifiable information is included, ensuring full anonymity and compliance with privacy standards.


\section{Results}

\subsection{Emotion Analysis}

\begin{figure}[ht]
    \centering
    \includegraphics[width=1\linewidth]{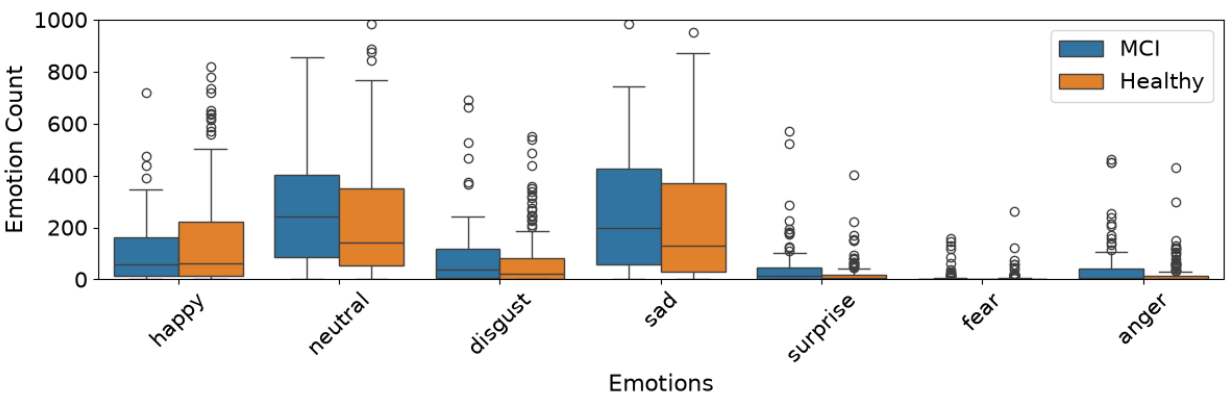}
    \caption{Relationship between Emotion Count and Emotion}
    \label{fig: Relationship between Emotion Count and Emotion}
\end{figure}

Figure~\ref{fig: Relationship between Emotion Count and Emotion} presents a boxplot illustrating the distribution of emotion counts across participants. The results indicate that individuals in the healthy group (orange) exhibit a higher frequency of positive emotions, such as happy and neutral, whereas participants with Mild Cognitive Impairment (MCI) (blue) display a greater occurrence of negative emotions, including sadness, disgust, and surprise. This trend suggests that emotional expressivity and valence patterns may serve as behavioral indicators of cognitive decline.


\begin{figure}[ht]
    \centering
    \includegraphics[width=1\linewidth]{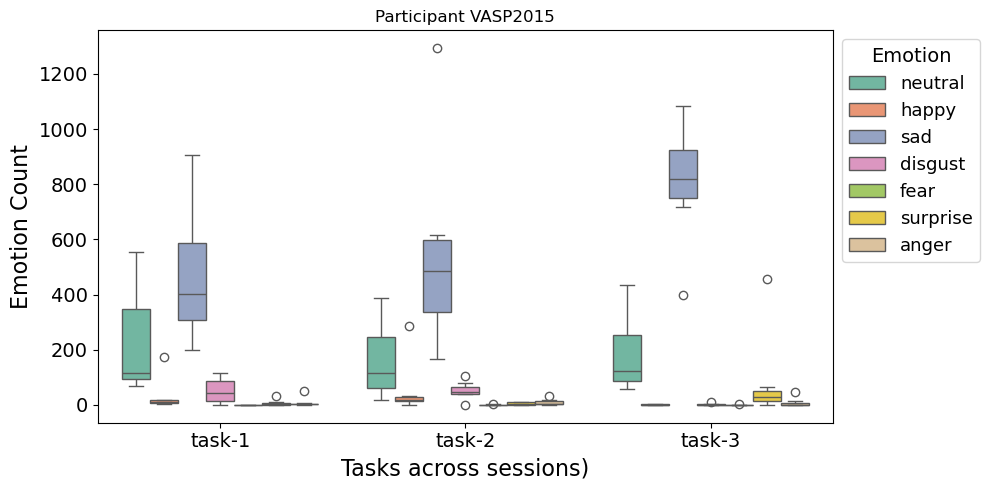}
    \caption{Emotions' distribution for a MCI Participant}
    \label{fig: Emotions across acts of a person with MCI}
\end{figure}

Figure~\ref{fig: Emotions across acts of a person with MCI} illustrates emotion variation across tasks for a participant with MCI. Negative affect, particularly sadness (blue), is prevalent across tasks, while happiness (orange) appears infrequently, suggesting reduced emotional variability and diminished positive affect. Notably, task 3 shows a marked increase in sadness, likely corresponding to moments when the participant was searching for or formulating an answer, potentially reflecting increased cognitive effort or task-related frustration. Overall, the emotional profile appears more constrained and skewed toward negative affect.

\begin{figure}[ht]
    \centering
    \includegraphics[width=1\linewidth]{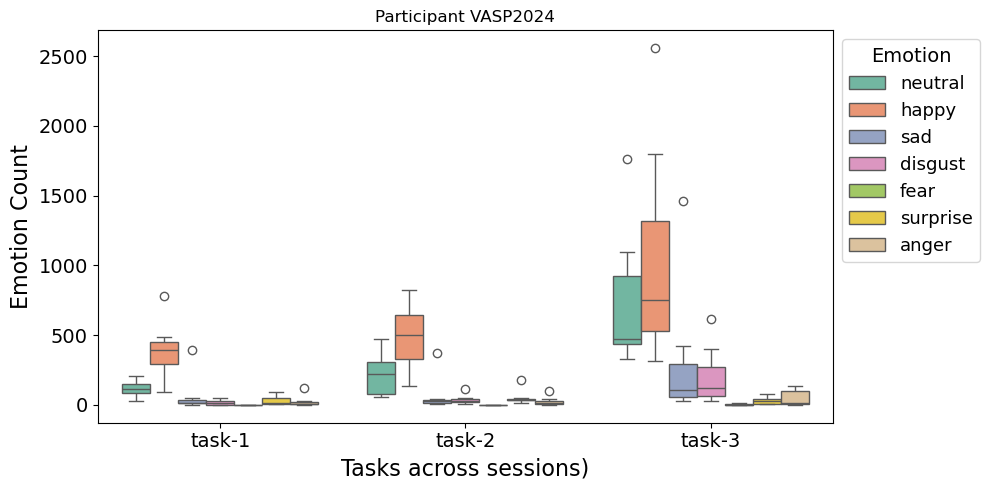}
    \caption{Emotions' distribution for a healthy Participant}
    \label{fig: Emotions across acts of a normal person}
\end{figure}

 In contrast, Figure~\ref{fig: Emotions across acts of a normal person} shows emotion variation for a cognitively healthy participant, where positive emotions, especially happiness, are more consistently expressed across tasks. This predominance of positive affect, along with broader emotional variability, may reflect stronger emotional regulation and preserved cognitive functioning. Even during more cognitively demanding tasks, the participant maintained positive emotional expression. In particular, task 3 exhibits richer emotional variation compared to the MCI participant, suggesting greater affective flexibility in response to task demands.

Task 1 shows a notable difference in how each participant initially perceives and engages with the task. As task complexity increases across tasks, the participant with MCI exhibits reduced emotional expressiveness and diminished variability in affective responses, whereas the healthy participant maintains richer emotional expression.

\subsection{Audio Analysis}

\begin{figure}[ht]
    \centering
    \includegraphics[width=1\linewidth]{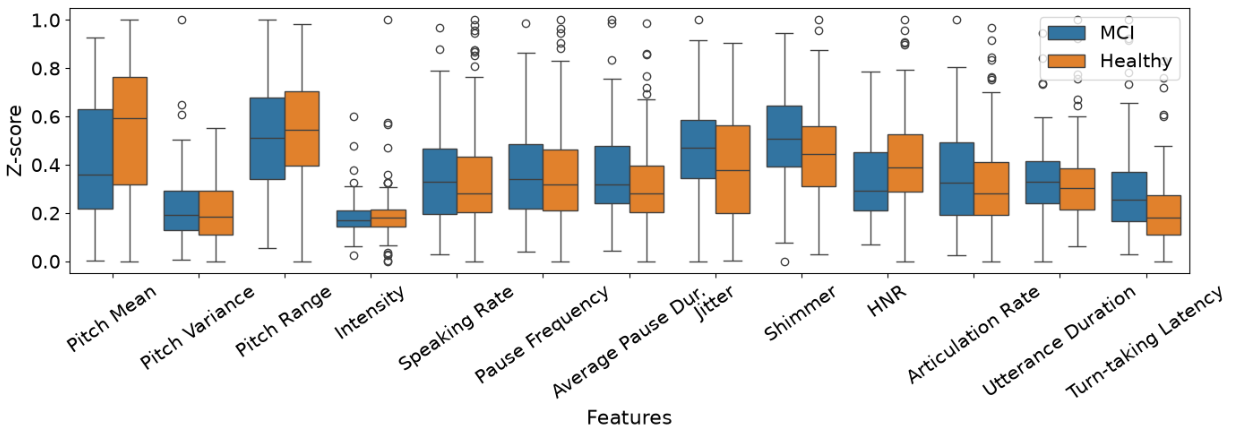}
    \caption{Audio Features Distributions by Cognitive Status}
    \label{fig: Normalized_Audio_Features_Distributions_by_Cognitive_Status}
\end{figure}

The figure~\ref{fig: Normalized_Audio_Features_Distributions_by_Cognitive_Status} illustrates the normalized distributions of key acoustic features for speakers with Mild Cognitive Impairment (MCI) and cognitively healthy controls. Individuals with MCI show lower pitch variability, reduced speaking and articulation rates, and longer pauses, indicating less dynamic prosody and slower, more effortful speech production. Measures of voice stability (jitter, shimmer, HNR) also suggest less stable phonation in the MCI group. Together, these patterns highlight consistent differences across prosodic, temporal, and voice-quality features, supporting their potential relevance as markers of early cognitive decline.

\begin{figure}[ht]
    \centering
    \includegraphics[width=1\linewidth]{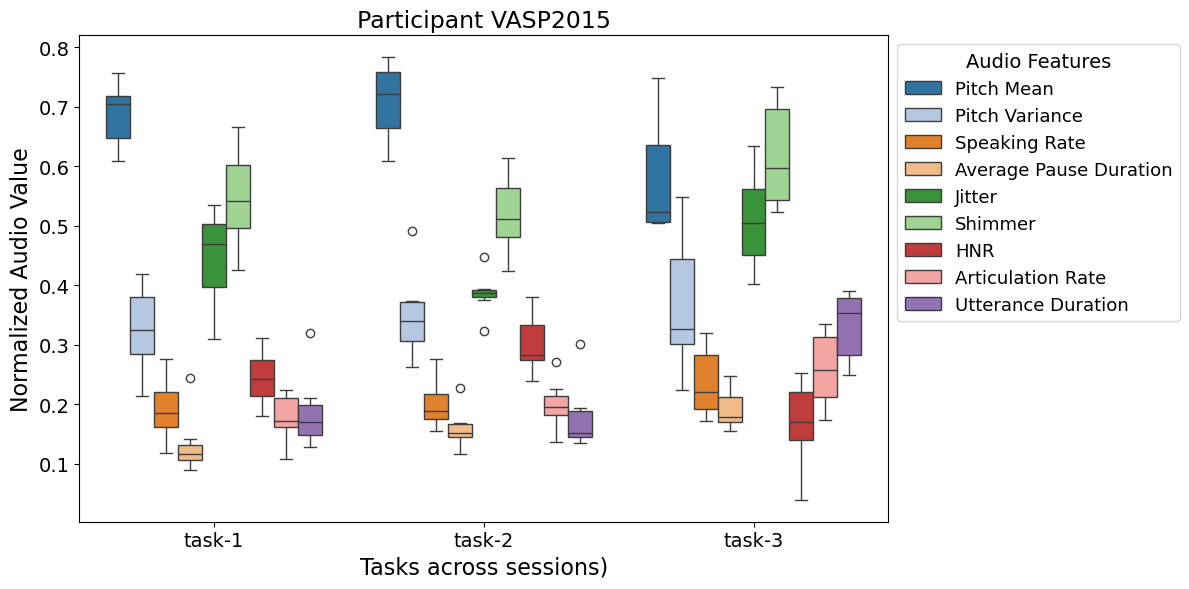}
    \caption{Audio Feature distribution for a MCI Participant}
    \label{fig: AudioFDistP2015}
\end{figure}

Figure \ref{fig: AudioFDistP2015} illustrates the distribution of some normalized acoustic features across sessions, enabling comparison of different tasks for participant VASP2015 with MCI. In task-1, the participant exhibits elevated pitch measures and moderate vocal instability, suggesting increased cognitive load during task initiation. Further, task-2 reveals a partial normalization of pitch but continued irregularity in temporal features such as speaking rate and articulation rate, indicating difficulty maintaining consistent speech rhythm during sustained interaction. By task-3, pitch variability increases again and hesitation markers, including pauses become more pronounced, maybe because of potential fatigue or diminished cognitive control as the tasks progress.

\begin{figure}[ht]
    \centering
    \includegraphics[width=1\linewidth]{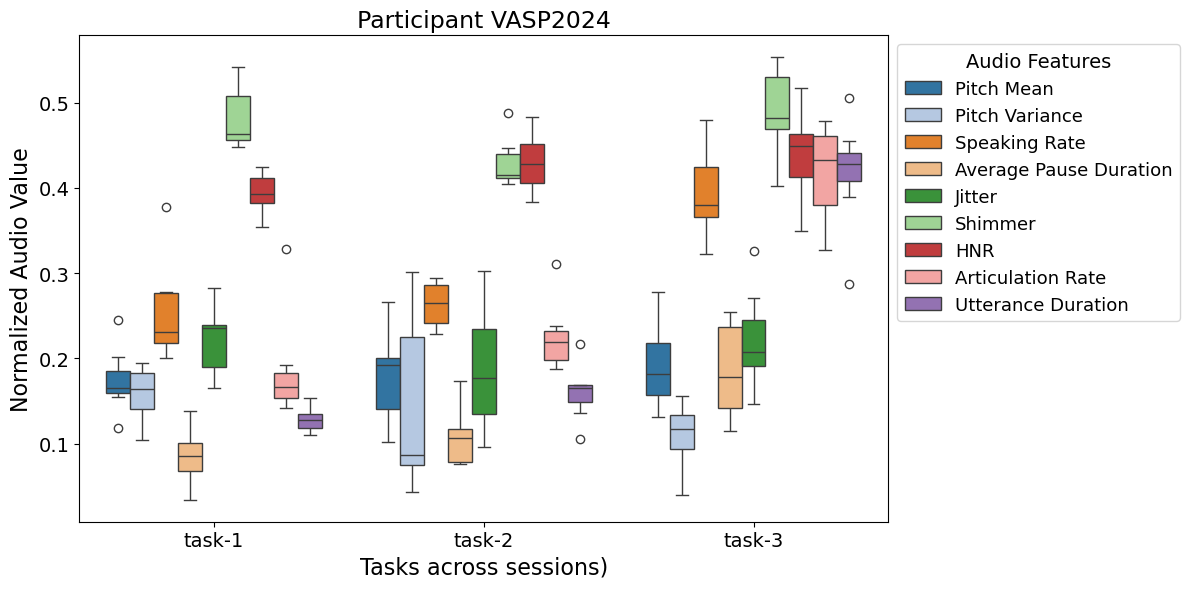}
    \caption{Audio Feature distribution for a healthy Participant}
    \label{fig: AudioFDistP2024}
\end{figure}

Figure \ref{fig: AudioFDistP2024} displays the same acoustic features for Participant VASP2024, a cognitively healthy control. This participant exhibits more compact and consistent distributions across nearly all measures. Task-1 shows controlled pitch values and balanced pause behavior, consistent with efficient task engagement from the outset. Task-2 maintains this stability, with articulation and speaking rates remaining tightly clustered and voice-quality measures (jitter, shimmer, HNR) indicating sustained phonatory control. Task-3 introduces more expressive variation such as slightly increased pause frequency and wider pitch range, but these shifts remain within the bounds of healthy adaptive modulation rather than cognitive strain. 

Utterance duration during task 3 suggests that this task elicits longer verbal responses for both groups. However, the healthy participant exhibits more sustained and consistently longer utterances, particularly in task 3, with tighter distributions indicating more stable response patterns. In contrast, the cognitively impaired participant shows greater variability in utterance duration, especially in the later task, suggesting less consistent speech production.
The cognitively impaired participant (VASP2015) generally exhibits higher variability in jitter and shimmer, particularly in tasks 2 and 3. Since jitter captures cycle-to-cycle frequency instability and shimmer reflects amplitude instability, elevated or more dispersed values may indicate less stable phonation and greater vocal irregularity. 
 In contrast, VASP2024 exhibits controlled, stable, and predictable acoustic behavior across sessions. The contrast between the two distributions supports evidence that MCI is associated with increased acoustic variability, disrupted speech timing, and reduced vocal stability, highlighting the potential of speech-based measures as sensitive indicators for early cognitive decline. Moreover, the MCI participant displays task-dependent destabilization, with greater variability and performance decline from task-1 to task-3, while the healthy participant maintains consistent control with slight changes.

\subsection{Model Analysis}

\begin{figure}[ht]
    \centering
    \includegraphics[width=1\linewidth]{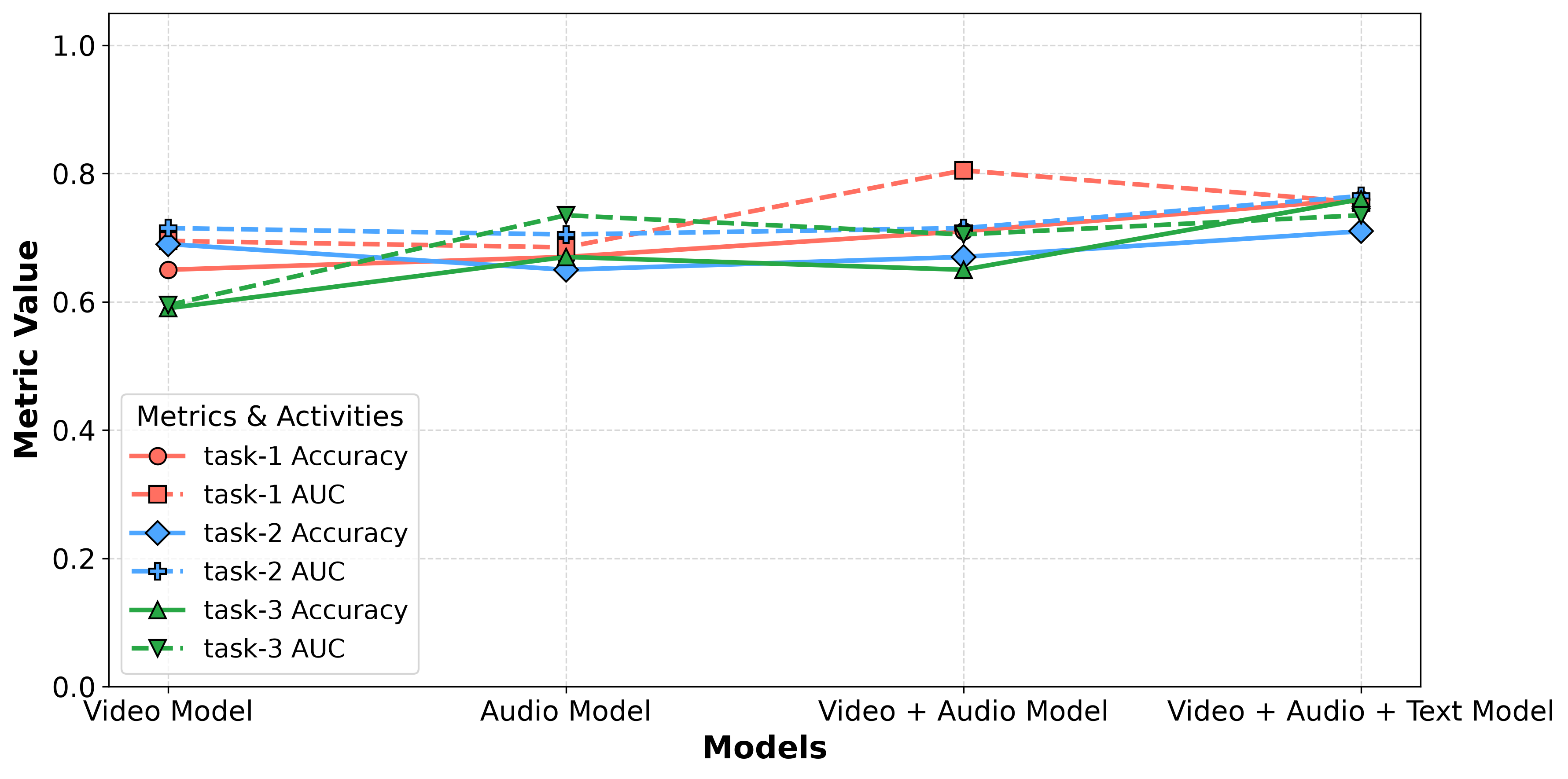}
    \caption{Comparison of Accuracy and AUC across tasks}
    \label{fig: COMP Acc and AUC}
\end{figure}

Figure ~\ref{fig: COMP Acc and AUC} presents a comparative analysis of four models across three experimental tasks. The plot displays Accuracy (solid lines with circular, diamond, and triangular markers) and AUC (dashed lines with square, plus, and inverted triangle markers) for each task. 
The Video model performs particularly well on task-1 and task-2, suggesting that these tasks provide richer and more discriminative facial expressions associated with cognitive state. In contrast, the Audio model achieves stronger performance on task-3, indicating that question answering better exposes verbal inconsistencies and speech-based cognitive markers.
Notably, the Video + Audio + Text model demonstrates the most consistent performance across all tasks, maintaining relatively stable Accuracy and AUC values.
Across models, tasks 2 and 3 consistently provide more informative signals, reflected in higher Accuracy and AUC scores, whereas task-1 appears comparatively less discriminative, exhibiting slightly lower performance and greater variability. 

\begin{figure}[ht]
    \centering
    \includegraphics[width=1\linewidth]{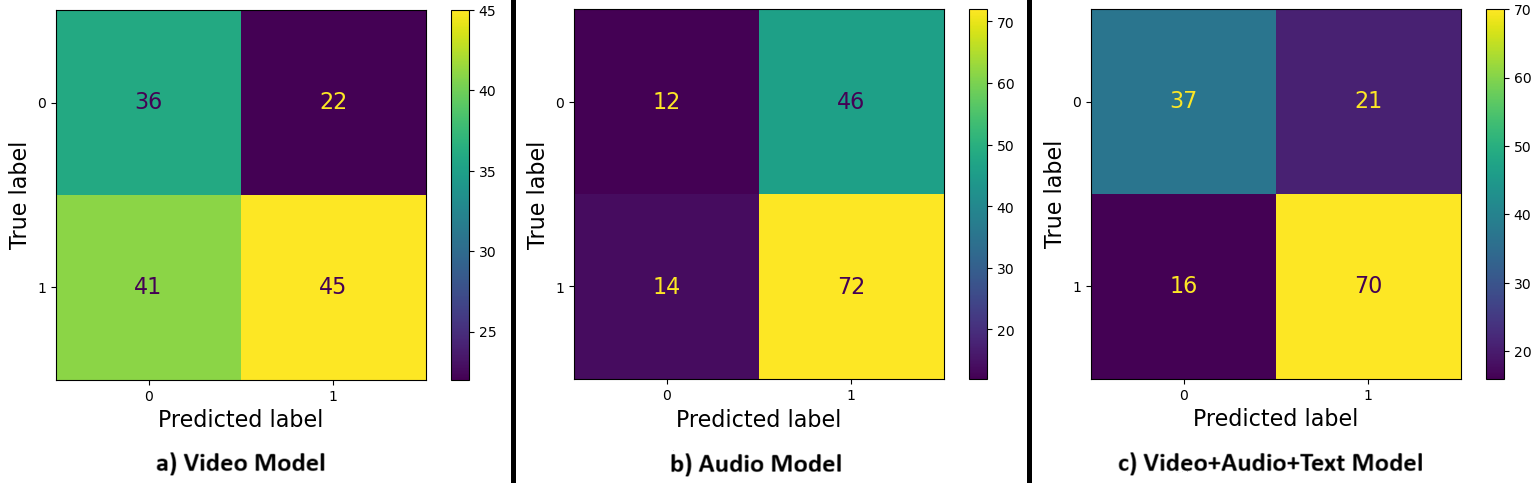}
    \caption{Comparison across different models}
    \label{fig: Emotion-Model- CM}
\end{figure}

Figure~\ref{fig: Emotion-Model- CM} illustrates how classification errors evolve as additional modalities are incorporated (0 = MCI). For this analysis, each task was treated as an independent data point, enabling an across-task rather than within-task analysis. This allows the model to capture variations in behavioral and emotional patterns across different tasks. The unimodal models exhibit a tendency toward dominant-class bias, leading to asymmetric misclassification patterns. 
The Video Model provides a baseline performance, correctly identifying 36 MCI participants, but suffers from substantial false negatives, misclassifying 41 healthy participants as MCI, indicating limited discriminative capability when relying on visual emotional cues alone. The Audio Model improves class separation by reducing false negatives from 41 to 14 and substantially increasing correct healthy classifications (72), though it introduces a high number of false positives for the MCI class, suggesting a bias toward predicting healthy controls. 
The combined Video+Audio+Text model achieves the most balanced and reliable detection, correctly identifying 37 MCI participants while reducing overall misclassifications to 16 false negatives and 21 false positives, demonstrating that integrating complementary behavioral signals mitigates dominant-class bias and improves class separability.
\section{Discussion}
The findings of this study demonstrate that natural conversational behavior contains meaningful signals for the early detection of cognitive impairment. Facial emotion patterns and AU dynamics reveal clear distinctions between cognitively healthy and MCI participants, with the latter exhibiting a higher prevalence of negative affect and reduced emotional variability. In addition, audio-based biomarkers and linguistic features provide complementary indicators of impairment, capturing disruptions in prosody, speech fluency, and semantic coherence. Together, these results highlight the importance of integrating visual, acoustic, and linguistic cues to obtain a more comprehensive characterization of early cognitive decline in naturalistic interaction settings.
Task-wise comparisons showed that tasks requiring question-generation or spontaneous responses produced more informative video signals, while predefined question answering provided more informative audio signals.
Looking ahead, several directions can strengthen and extend this work. Expanding the dataset with more participants and diverse demographics will improve generalizability.
We will also focus on longitudinal and cross-session modeling using Transformer-based sequence models to capture subtle behavioral changes across sessions.
Additionally, we will focus on improving multimodal fusion strategies and exploring LLMs to provide more interpretable and clinically meaningful reasoning over behavioral signals.

\section{Conclusion}
This work introduces a multimodal preprocessing pipeline that integrates facial behavior, audio biomarkers, and linguistic features to distinguish cognitively healthy individuals from those with MCI.
In contrast to traditional models, our framework leverages the inherently multimodal nature of human communication to capture a richer and more comprehensive representation of cognitive function.
By analyzing emotional dynamics, we show that meaningful cognitive signals emerge even during casual conversational tasks, with Command-Reading and Command-Generation revealing richer behavioral markers than structured question-answering tasks. Moreover, acoustic and linguistic cues provide complementary information, particularly in tasks such as Command-Generation and Question-Answering, highlighting the value of modeling task-dependent multimodal behavior across tasks.
Our results demonstrate that combining all three modalities substantially improves predictive performance (accuracy = 0.75, AUC = 0.80), outperforming any unimodal system. 
This approach enables scalable, continuous monitoring through natural interactions without clinical instrumentation, reflecting a real-world environment. 
This framework can be embedded into telehealth workflows, smart-home devices, video-calling platforms, and digital wellness applications, enabling continuous, remote monitoring and early alerts for cognitive decline.

\section{Acknowledgement}
This research was funded by the US National Institutes of Health National Institute on Aging R01AG067416.

\bibliographystyle{plain}

\bibliography{references}

@article{AudioTrans1,
  title   = {Multimodal Siamese networks for dementia detection from speech in women},
  author  = {Ksibi, A. and Walha, A. and Zakariah, M. and et al.},
  journal = {Scientific Reports},
  volume  = {15},
  year    = {2025},
  doi     = {10.1038/s41598-025-13902-7}
}

@article{Face1,
  title   = {Deep learning inspired game-based cognitive assessment for early dementia detection},
  author  = {Kundu Maji, P. and Acharya, S. and Paul, P. and et al.},
  journal = {Engineering Applications of Artificial Intelligence},
  volume  = {142},
  pages   = {109901},
  year    = {2025},
  doi     = {10.1016/j.engappai.2024.109901}
}

@article{Speech1,
  title   = {Towards Computer-Based Automated Screening of Dementia Through Spontaneous Speech},
  author  = {Chlasta, K. and Wołk, K.},
  journal = {Frontiers in Psychology},
  volume  = {11},
  year    = {2021},
  doi     = {10.3389/fpsyg.2020.623237}
}

@article{Speech2,
  title   = {Early dementia detection with speech analysis and machine learning techniques},
  author  = {Jahan, Z. and Bhatia, S. and Saraee, M.},
  journal = {Discover Sustainability},
  volume  = {5},
  year    = {2024},
  doi     = {10.1007/s43621-024-00217-2}
}

@article{brain1,
  title   = {Evaluation of MRI-based machine learning approaches for computer-aided diagnosis of dementia in a clinical data warehouse},
  author  = {Bottani, S. and Burgos, N. and Maire, A. and et al.},
  journal = {Medical Image Analysis},
  volume  = {89},
  pages   = {102903},
  year    = {2023},
  doi     = {10.1016/j.media.2023.102903}
}

@article{brain2,
  title   = {Differential dementia detection from multimodal brain images in a real-world dataset},
  author  = {Leming, M. and Im, H.},
  journal = {Alzheimer's \& Dementia},
  volume  = {21},
  number  = {7},
  pages   = {e70362},
  year    = {2025},
  doi     = {10.1002/alz.70362}
}

@misc{alzheimers_mci,
  author       = {{U.S. Department of Health and Human Services}},
  title        = {Mild Cognitive Impairment},
  year         = {2025},
  url          = {https://www.alzheimers.gov/alzheimers-dementias/mild-cognitive-impairment},
  note         = {[Online; accessed 7-October-2025]},
  organization = {Alzheimers.gov}
}

@article{Sent1,
  title   = {Adaptive multimodal transformer based on exchanging for multimodal sentiment analysis},
  author  = {Tuerhong, G. and Fu, F. and Wushouer, M.},
  journal = {Scientific Reports},
  volume  = {15},
  pages   = {27265},
  year    = {2025},
  doi     = {10.1038/s41598-025-11848-4}
}

@article{Sent2,
  title   = {Multimodal sentiment analysis based on multi-layer feature fusion and multi-task learning},
  author  = {Cai, Y. and Li, X. and Zhang, Y. and et al.},
  journal = {Scientific Reports},
  volume  = {15},
  pages   = {2126},
  year    = {2025},
  doi     = {10.1038/s41598-025-85859-6}
}

@article{LLMH1,
  title   = {Large Language Models in Healthcare and Medical Applications: A Review},
  author  = {Maity, S. and Saikia, M. J.},
  journal = {Bioengineering},
  volume  = {12},
  number  = {6},
  pages   = {631},
  year    = {2025},
  doi     = {10.3390/bioengineering12060631}
}

@article{AudioAn1,
  title   = {Voiceprints of cognitive impairment: analyzing digital voice for early detection of Alzheimer’s and related dementias},
  author  = {Rezaii, N. and Wong, B. and Aisen, P. and et al.},
  journal = {npj Dementia},
  volume  = {1},
  pages   = {35},
  year    = {2025},
  doi     = {10.1038/s44400-025-00040-0}
}

@article{AudioAn2,
  title   = {Detection of Mild Cognitive Impairment From Non-Semantic, Acoustic Voice Features: The Framingham Heart Study},
  author  = {Ding, H. and Lister, A. and Karjadi, C. and et al.},
  journal = {JMIR Aging},
  volume  = {7},
  pages   = {e55126},
  year    = {2024},
  doi     = {10.2196/55126}
}

@article{MiVOLO,
  title   = {MiVOLO: Multi‑input Transformer for Age and Gender Estimation},
  author  = {Kuprashevich, M. and Tolstykh, I.},
  journal = {arXiv preprint},
  eprint  = {arXiv:2307.04616},
  year    = {2023},
  doi     = {10.48550/arXiv.2307.04616}
}

@article{ResEmoteNet,
  title   = {ResEmoteNet: Bridging Accuracy and Loss Reduction in Facial Emotion Recognition},
  author  = {Roy, A. K. and Kathania, H. K. and Sharma, A. and et al.},
  journal = {arXiv preprint},
  eprint  = {arXiv:2409.10545},
  year    = {2024},
  doi     = {10.48550/arXiv.2409.10545}
}

@misc{CogTiPRO,
author = {Qi, Kristin and Zhu, Youxiang and Summerour, Caroline and et al},
year = {2025},
month = {05},
pages = {},
title = {Cog-TiPRO: Iterative Prompt Refinement with LLMs to Detect Cognitive Decline via Longitudinal Voice Assistant Commands},
doi = {10.48550/arXiv.2505.17137}
}

@inproceedings{OpenFace1,
  title     = {OpenFace: An open source facial behavior analysis toolkit},
  author    = {Baltrušaitis, T. and Robinson, P. and Morency, L.-P.},
  booktitle = {2016 IEEE Winter Conference on Applications of Computer Vision (WACV)},
  pages     = {1-10},
  year      = {2016},
  doi       = {10.1109/WACV.2016.7477553}
}

@article{text1,
  title   = {Contrastive text embeddings with effective sample mining for enhanced disease diagnosis},
  author  = {Zhang, X. and Zhao, G. and Ren, Y. and et al.},
  journal = {Array},
  volume  = {28},
  pages   = {100578},
  year    = {2025},
  doi     = {10.1016/j.array.2025.100578}
}

@article{text2,
  title   = {Med-BERT: pretrained contextualized embeddings on large-scale structured electronic health records for disease prediction},
  author  = {Rasmy, L. and Xiang, Y. and Xie, Z. and et al.},
  journal = {npj Digital Medicine},
  volume  = {4},
  pages   = {86},
  year    = {2021},
  doi     = {10.1038/s41746-021-00455-y}
}

@article{TS1,
  title   = {Time-series visual explainability for Alzheimer’s disease progression detection for smart healthcare},
  author  = {Rahim, N. and Abuhmed, T. and Mirjalili, S. and et al.},
  journal = {Alexandria Engineering Journal},
  volume  = {82},
  pages   = {484-502},
  year    = {2023},
  doi     = {10.1016/j.aej.2023.09.050}
}

@article{TS2,
  title   = {Dementia risk analysis using temporal event modeling on a large real-world dataset},
  author  = {Taylor, R. A. and Gilson, A. and Chi, L. and et al.},
  journal = {Scientific Reports},
  volume  = {13},
  pages   = {22618},
  year    = {2023},
  doi     = {10.1038/s41598-023-49330-8}
}

@INPROCEEDINGS{RI,
  author={Singh, Rishank and Vaish, Sankalp and Liang, Xiaohui and et al.},
  booktitle={2024 IEEE International Conference on E-health Networking, Application \& Services (HealthCom)}, 
  title={Clinical Geriatric Functional Assessment Using Wearable Sensing and Machine Learning}, 
  year={2024},
  volume={},
  number={},
  pages={1-7},
  doi={10.1109/HealthCom60970.2024.10880842}}

@INPROCEEDINGS{RI2,
  author={Kurtz, Eli and Zhu, Youxiang and Driesse, Tiffany and et al},
  booktitle={ICASSP 2023 - 2023 IEEE International Conference on Acoustics, Speech and Signal Processing (ICASSP)}, 
  title={Early Detection of Cognitive Decline Using Voice Assistant Commands}, 
  year={2023},
  volume={},
  number={},
  pages={1-5},
  month={June},}

@inproceedings{RI3,
  title     = {Analyzing Multimodal Features of Spontaneous Voice Assistant Commands for Mild Cognitive Impairment Detection},
  author    = {Lin, N. and Zhu, Y. and Liang, X. and et al.},
  booktitle = {Proc. Interspeech},
  pages     = {3030--3034},
  year      = {2024},
  doi       = {10.21437/interspeech.2024-2288}
}

@article{AU1,
  title   = {Explaining facial action units' correlation with hypomimia and clinical scores in Parkinson’s disease},
  author  = {Filali Razzouki, A. and Jeancolas, L. and Sambin, S. and et al.},
  journal = {npj Parkinson’s Disease},
  volume  = {11},
  pages   = {53},
  year    = {2025},
  doi     = {10.1038/s41531-025-00895-3}
}
\end{document}